\documentclass[]{article}
\usepackage{authblk}

\usepackage{fullpage}
\usepackage{pifont}
\newcommand{\cmark}{\text{\ding{51}}}%
\newcommand{\xmark}{\text{\ding{55}}}%
\usepackage[utf8]{inputenc} 
\usepackage[T1]{fontenc}    
\usepackage{hyperref}       
\usepackage{url}            
\usepackage{booktabs}       
\usepackage{amsfonts}       
\usepackage{nicefrac}       
\usepackage{microtype}      
\usepackage{graphicx}
\usepackage[export]{adjustbox}
\usepackage{subfig}
\usepackage{stackengine}
\usepackage{mathtools}
\usepackage{multirow}
\usepackage{xcolor}
\usepackage{Definitions}
\usepackage{enumitem}
\usepackage{tikz}
\usepackage{algorithmic}
\usepackage{hhline}
\usepackage{marginnote}
\usepackage{setspace}
\usepackage[ruled,noline,norelsize]{algorithm2e}

\newcommand{\badge}[1]{\ifmmode\text{\texttt{#1}}\else\texttt{#1}\fi}
\newcommand{\xhdr}[1]{\vspace{1.2mm}\noindent{{\bf #1.}}}

\newcommand{\explain}[2]{\underset{\mathclap{\overset{\uparrow}{#2}}}{#1}}
\newcommand{\explainup}[2]{\overset{\mathclap{\underset{\downarrow}{#2}}}{#1}}
\newcommand*{\defeq}{\coloneqq}
\newcommand{\icol}[1]{
  \bigl[\begin{smallmatrix}#1\end{smallmatrix}\bigr]%
}
\newcommand\stackequal[2]{\mathrel{\stackunder[2pt]{\stackon{=}{$\scriptscriptstyle#1$}}{$\scriptscriptstyle#2$}}}

\renewcommand{\algorithmicrequire}{\textbf{Input:}}
\renewcommand{\algorithmicensure}{\textbf{Output:}}
\newcommand{\algorithmicreturn}{\textbf{return }}
\renewcommand{\algorithmiccomment}[1]{\hfill // #1}

\newcommand{\forget}{\begin{tikzpicture} \draw[red!60!white,thick] (-0.08,-0.08) -- (0.08,0.08);\draw[red!60!white,thick] (0.08,-0.08) -- (-0.08,0.08);\end{tikzpicture}}
\newcommand{\recall}{\begin{tikzpicture}\fill[green!60!white] (0,0) circle (0.097cm);\end{tikzpicture}}

\newcommand{\memorize}{{\textsc{Memorize}}}

\graphicspath{{./FIG/}}

\title{Optimizing Human Learning}

\author[1,2]{Behzad Tabibian}
\author[1]{Utkarsh Upadhyay}
\author[1]{Abir De}
\author[3]{Ali Zarezade}
\author[2]{Bernhard Sch\"{o}lkopf}
\author[1]{\mbox{Manuel Gomez-Rodriguez}}

\affil[1]{MPI for Software Systems, utkarshu@mpi-sws.org, ade@mpi-sws.org, manuelgr@mpi-sws.org}
\affil[2]{MPI for Intelligent Systems Systems, btabibian@tue.mpg.de, bs@tue.mpg.de}
\affil[3]{Sharif University, zarezade@ce.sharif.edu}

\date{}

\begin{document}


\maketitle

\begin{abstract}
Spaced repetition is a technique for efficient memorization which uses repeated, spaced review of content to improve long-term 
retention.
Can we find the optimal reviewing schedule to maximize the benefits of spaced repetition?
In this paper, we introduce a novel, flexible representation of spaced repetition using the framework of marked temporal point processes 
and then address the above question as an optimal control problem for stochastic differential equations with jumps.
For two well-known human memory models, we show that the optimal reviewing schedule is given by the recall probability of the content 
to be learned. As a result, we can then develop a simple, scalable online algorithm, \memorize, to sample the optimal reviewing times.
Experiments on both synthetic and real data gathered from Duolingo, a popular language-learning online platform, 
show that our algorithm may be able to help learners memorize more effectively than alternatives.
\end{abstract}

\section{Introduction}
\label{sec:introduction}
Our ability to remember a piece of information depends critically on the number of times we have reviewed it 
and the time elapsed since the last review, as first shown by a seminal study by Ebbinghaus~\cite{ebbinghaus1885memory}.
The effect of these two factors have been extensively investigated in the experimental psychology literature~\cite{dempster1989spacing, melton1970situation}, particularly in
second language acquisition research~\cite{atkinson1972optimizing, bloom1981effects, cepeda2006distributed, pavlik2008using}.
Moreover, these empirical stu\-dies have motivated the use of \emph{flashcards}, small pieces of information a learner repeatedly reviews following a sche\-dule determined by a \emph{spaced
repetition} algorithm~\cite{sr}, whose goal is to ensure that learners spend more (less) time working on forgotten (recalled) information.

In recent years, spaced repetition software and online platforms such as Mnemosyne\footnote{\scriptsize http://mnemosyne-proj.org/}, Synap\footnote{\scriptsize http://www.synap.ac},
SuperMemo\footnote{\scriptsize https://www.supermemo.com}, or Duolingo\footnote{\scriptsize http://www.duolingo.com} have become increasingly popular, often re\-pla\-cing the use of
physical flashcards.
The promise of these softwares and online platforms is that automated fine-grained mo\-ni\-to\-ring and greater degree of control will result in more effective spaced
repetition algorithms.
However, most of these algorithms are simple rule-based heuristic with a few hard-coded parameters~\cite{sr}---principled data-driven models and algorithms with provable
guarantees have been largely missing until very recently~\cite{novikoff2012education, reddy2016unbounded}.
Among these recent notable exceptions, the work most closely related to ours is by Reddy et al.~\cite{reddy2016unbounded}, who proposed a queueing network model for
a particular spaced repetition method---the Leitner system~\cite{leitner1974} for reviewing flashcards---and then developed a heuristic approximation for scheduling
reviews.
However, their heuristic does not have provable guarantees, it does not adapt to the learner'{}s performance over time, and it is specifically designed for Leitner systems.

In this paper, we first introduce a novel, flexible representation of spaced repetition using the framework of marked temporal point processes~\cite{AalBorGje08}.
For two well-known human memory models, we use this presentation to express the dynamics of a learner'{}s forgetting rates and recall probabilities for
the content to be learned by means of a set of stochastic differential equations (SDEs) with jumps.
Then, we can find the optimal reviewing schedule for spaced repetition by solving a stochastic optimal control problem for SDEs with jumps~\cite{hanson2007, cheshire2017zarezade, redqueen17wsdm}.
In doing so, we need to introduce a proof technique of independent interest (refer to Appendices~\ref{app:pro-hjb-proposals} and~\ref{app:optimal-intensity}).

The solution uncovers a linear relationship between the optimal reviewing intensity and the recall probability of the content to be learned, which
allows for a simple, scalable online algorithm, which we name \memorize, to sample the optimal reviewing times (Algorithm~\ref{alg:memorize}).
Finally, we experiment with both synthetic and real data gathered from Duolingo, a popular language-learning online platform,
and show that our algorithm may be able to help learners memorize more effectively than alternatives.
To facilitate research in this area within the machine learning community, we are releasing an implementation of our algorithm at \href{http://learning.mpi-sws.org/memorize/}{http://learning.mpi-sws.org/memorize/}.

\xhdr{Further related work}
There is a rich literature which tries to ascertain which model of human memory predicts performance best~\cite{settlestrainable,wixted2007wickelgren}.
Our aim in this work is to provide a methodology to derive an optimal reviewing schedule given a choice of human memory model. 
Hence, we apply our methodology to two of the most popular memory models from the literature---exponential and power-law forgetting curve models.

The task of designing reviewing schedules also has a rich history, starting with the Leitner system itself~\cite{leitner1974}.
In this context, Metzler-Baddeley et al.~\cite{metzler2009does} have recently shown that adaptive reviewing schedules perform better than non-adaptive ones using data from
SuperMemo.
In doing so, they proposed an algorithm that schedules reviews just as the learner is \emph{about} to forget an \emph{item}, \ie, when the probability of recall falls below a 
threshold.
Lindsey et al.~\cite{lindsey2014improving} have also used a similar idea for scheduling reviews, albeit with a model of recall inspired by ACT-R and Multiscale Context
Model~\cite{pashler2009predicting}.
In this work, we use such heuristic, which does not have theoretical guarantees, as a baseline (``Threshold'') in our experiments.

Finally, another line of research has pursued locally optimal scheduling by identifying which item would benefit the most from a review.
Pavlik et al.~\cite{pavlik2008using} have used the ACT-R model to make locally optimal decisions about which item to review by greedily selecting the item which is closest
to its maximum \emph{learning rate} as a heuristic.
Mettler et al.~\cite{mettler2016comparison} have also employed a similar heuristic (ARTS system) to arrive at a reviewing schedule by taking response time into account.
In this work, our goal is devising strategies which are globally optimal and allow for explicit bounds on the rate of reviewing.



\section{Problem Formulation}
\label{sec:formulation}
In this section, we first briefly revisit two popular memory models we will use in our work.
Then, we describe how to represent spaced repetition using the framework of marked temporal point processes.
Finally, we conclude with a statement of the spaced repetition problem.

\xhdr{Modeling human memory} Following previous work in the psychology literature~\cite{averell2011form, ebbinghaus1885memory, loftus1985evaluating, wixted2007wickelgren},
we consider the exponential and the power-law forgetting curve models with binary recalls (\ie, a user either completely recalls or forgets an item).

The probability of recalling item $i$ at time $t$ for the exponential forgetting curve model is given by
\begin{equation} \label{eq:recall-probability}
m_i(t) := \PP(r_i(t) = 1) = \exp\left(-n_i(t) (t - t_r)\right),
\end{equation}
where $t_r$ is the time of the last review and $n_i(t) \in \RR^{+}$ is the forgetting rate\footnote{\scriptsize Previous works often use the inverse of the forgetting rate,
referred as memory strength or half-life, $s(t) = n^{-1}(t)$~\cite{reddy2016unbounded, settlestrainable}. However, it will be more tractable for us to work with the
forgetting rates.} at time $t$, which may depend on many factors, \eg, item difficulty and number of previous (un)successful recalls of the item.
The probability of recalling item $i$ at time $t$ for the power-law forgetting curve model is given by
\begin{equation} \label{eq:recall-probability-pl}
m_i(t) := \PP(r_i(t) = 1) = (1 + \omega (t - t_r) )^{-n_i(t)},
\end{equation}
where $t_r$ is the time of the last review, $n_i(t) \in \RR^{+}$ is the forgetting rate and $\omega$ is a time scale parameter.
%
%
Remarkably, despite their simplicity, the above functional forms have been recently shown to provide accurate quantitative predictions at a user-item level in large scale web
data~\cite{reddy2016unbounded, settlestrainable}.

In the remainder of the paper, for ease of exposition, we derive the optimal reviewing schedule and report experimental results only for the exponential forgetting curve model. Appendix~\ref{app:power-law-derivations}
contains the derivation of the optimal reviewing schedule for the power-law forgetting curve model as well as its experimental validation. Other more complex models of memory can also be expressed using SDEs, \eg,
see Appendix~\ref{app:other-sde} for the MCM model of memory~\cite{pashler2009predicting}.
In this paper, we find a solution to the optimization problem for the simpler exponential and powerlaw models of memory, leaving optimization with more complex models for future work.

\xhdr{Modeling spaced repetition} Given a learner who wants to memorize a set of items $\Ical$ using spaced repetition, \ie, repeated, spaced
reviews of the items, we represent each reviewing event as a triplet
\begin{equation*}
e~~\defeq~~(~\explain{i}{\text{item}},~~~\explainup{t}{\text{time}},~~~\explain{r}{\text{recall}}~),
\end{equation*}
which means that the learner reviewed item $i \in \Ical$ at time $t$ and either recalled it ($r = 1)$ or forgot it ($r = 0$).
Here, note that each reviewing event includes the outcome of a test (\ie, a recall) and this is a key difference from the paradigm used by several laboratory studies~\cite{cepeda2006distributed,cepeda2008spacing},
which consider a sequence of reviewing events followed by a single test.
In other words, our data consists of \textsc{test}/\textsc{review}-\dots-\textsc{test}/\textsc{review} sequences, in contrast, the data in those studies consists
of \textsc{review}-\ldots-\textsc{review}-\textsc{test} sequences\footnote{\scriptsize In most spaced repetition software and online platforms such as Mnemosyne,
Synap, or Duolingo, the learner is tested in each review, \ie, the learner follows \textsc{test}/\textsc{review}-\dots-\textsc{test}/\textsc{review} sequences.}.

In the above representation, we model the recall $r$ using the memory model defined by Eq.~\ref{eq:recall-probability}, \ie, $r \sim \text{Bernoulli}(m_i(t))$,
and we keep track of the reviewing times using a multidimensional counting process $\Nb(t)$, in which the $i$-th entry, $N_i(t)$, counts the
number of times the learner has reviewed item $i$ up to time $t$.
%
Following the literature on temporal point processes~\cite{AalBorGje08}, we characterize these counting processes using their corresponding intensities,
\ie, $\EE[d\Nb(t)] = \ub(t)dt$, and think of the recall $r$ as their binary \emph{marks}.
%
%
Moreover, every time a learner reviews an item, the recall $r$ has been experimentally shown to have an effect on the forgetting rate of the
item~\cite{dempster1989spacing, reddy2016unbounded, settlestrainable}.
In particular, using large scale web data from Duolingo, Settles et al.~\cite{settlestrainable} have provided strong empirical evidence that (un)successful recalls
of an item $i$ during a review have a multiplicative effect on the forgetting rate $n_i(t)$---a successful recall at time $t_r$ changes the forgetting rate
by $(1 - \alpha_i)$, \ie, $n_i(t) = (1 - \alpha_i) n_i(t_r)$, $\alpha_i \leq 1$, while an unsuccessful recall changes the forgetting rate by $(1 + \beta_i)$, \ie,
$n_i(t) = (1 + \beta_i) n_i(t_r)$, $\beta_i \geq 0$, where $\alpha_i$ and $\beta_i$ are item specific parameters which can be found using historical data.
In this context, the initial forgetting rate, $n_i(0)$, captures the difficulty of the item, with more difficult items having higher initial forgetting rates compared
to easier items.

Hence, we express the dynamics of the forgetting rate $n_i(t)$ for each item $i \in \Ical$ using the following stochastic differential equation (SDE) with
jumps:
\begin{equation} \label{eq:forgetting-rate}
dn_i(t) = -\alpha_i n_i(t) r_i(t) dN_i(t) + \beta_i n_i(t)(1-r_i(t)) dN_i(t),
\end{equation}
where $N_i(t)$ is the corresponding counting process and $r_i(t) \in \{0, 1\}$ indicates whether item $i$ has been successfully
recalled at time $t$.
Here, we would like to highlight that:
(i) the forgetting rate, as defined above, is a Markov process and this will be useful in the derivation of the optimal reviewing schedule;
(ii) the Leitner system~\cite{leitner1974} with exponential spacing can also be cast using this formulation with particular choices of $\alpha_i$
and $\beta_i$ and the same initial forgetting rate, $n_i(0)=n(0)$, for all items;
and, (iii) several laboratory studies, in which learners follow sequences \textsc{review}-\ldots-\textsc{review}-\textsc{test}, suggest the
pa\-ra\-me\-ters $\alpha_i$ and $\beta_i$ should be time-varying since the retention rate follows an \emph{inverted U-shape}~\cite{cepeda2008spacing},
however, we found that in our dataset, in which learners follow sequences \textsc{test}/\textsc{review}-\dots-\textsc{test}/\textsc{review}, considering constant
$\alpha_i$ and $\beta_i$ is a valid approximation (refer to Appendix~\ref{app:reviewTimes}).

Given the above definition, one can also express the dynamics of the recall probability $m_i(t)$, defined by Eq.~\ref{eq:recall-probability}, by means of a SDE
with jumps using the following Proposition (proven in Appendix~\ref{app:pro-recall-probability}):
\begin{proposition} \label{pro:recall-probability}
Given an item $i \in \Ical$ with reviewing intensity $u_i(t)$, the recall probability $m_i(t)$, defined by Eq.~\ref{eq:recall-probability}, is a Markov process
whose dynamics can be defined by the following SDE with jumps:
\begin{equation} \label{eq:recall-probability-m}
dm_i(t) = -n_i(t)m_i(t)dt + (1-m_i(t)) dN_i(t),
\end{equation}
where $N_i(t)$ is the counting process associated to the reviewing intensity $u_i(t)$.
\end{proposition}
%
%
Expressing the dynamics of the forgetting rates and recall probabilities as SDEs with jumps will be very useful for the design of our stochastic optimal control
algorithm for spaced repetition.

\xhdr{The spaced repetition problem} Given a set of items $\Ical$, our goal is to find the optimal item re\-vie\-wing intensities $\ub(t) = [u_i(t)]_{i \in \Ical}$
that minimize the expected value of a particular convex loss function $\ell(\mb(t), \nbb(t), \ub(t))$ of the recall probability of the items, $\mb(t) = [m_i(t)]_{i \in \Ical}$,
the forgetting rates, $\nbb(t) = [n_i(t)]_{i \in \Ical}$, and the intensities themselves, $\ub(t)$, over a time window $(t_0, t_f]$, \ie,
\begin{align} \label{eq:spaced-repetition-problem}
\underset{\ub(t_0,t_f]}{\text{minimize}} & ~~\EE_{(\Nb, \rb)(t_0,t_f]} \left[ \phi(\mb(t_f), \nbb(t_f)) +\int_{t_0}^{t_f} \ell(\mb(\tau),\nbb(\tau), \ub(\tau))d\tau \right] \nonumber\\
\text{subject to} & ~~\ub(t) \geq 0 ~ \forall t \in (t_0,t_f),
\end{align}
where $\ub(t_0,t_f]$ denotes the item reviewing intensities from $t_0$ to $t_f$, the expectation is taken over all possible realizations
of the associated counting processes and (item) recalls, denoted as $(\Nb, \rb)(t_0,t_f]$, the loss function is nonincreasing (nondecreasing)
with respect to the recall probabilities (forgetting rates and intensities) so that it rewards long-lasting learning while limiting the number of item reviews,
and $\phi(\mb(t_f), \nbb(t_f))$ is an arbitrary penalty function.
Finally, note that the forgetting rates $\nbb(t)$ and recall probabilities $\mb(t)$, defined by Eq.~\ref{eq:forgetting-rate} and Eq.~\ref{eq:recall-probability-m},
depend on the reviewing intensities $\ub(t)$ we aim to optimize since $\EE[d\Nb(t)] = \ub(t) dt$.

\section{The \memorize{} Algorithm}
\label{sec:method}
In this section, we tackle the spaced repetition problem defined by Eq.~\ref{eq:spaced-repetition-problem} from the perspective of stochastic optimal control of
jump SDEs~\cite{hanson2007}. More specifically, we first derive a solution to the problem considering only one item, provide an efficient practical implementation
of the solution, and then generalize it to the case of multiple items.

\xhdr{Optimizing for one item} Given an item $i$ with reviewing intensity $u_i(t) = u(t)$ and associated counting process $N_i(t) = N(t)$, recall outcome $r_i(t) = r(t)$,
recall probability $m_i(t) = m(t)$ and forgetting rate $n_i(t) = n(t)$, we can rewrite the spaced repetition problem defined by Eq.~\ref{eq:spaced-repetition-problem}
as:
\begin{align} \label{eq:spaced-repetition-problem-one-item}
\underset{u(t_0,t_f]}{\text{minimize}} & ~~\EE_{(N, r)(t_0,t_f]} \left[ \phi(m(t_f), n(t_f))+\int_{t_0}^{t_f} \ell(m(\tau),n(\tau), u(\tau))d\tau \right] \nonumber\\
\text{subject to} & ~~u(t) \geq 0 ~ \forall t \in (t_0,t_f),
\end{align}
where, using Eq.~\ref{eq:forgetting-rate} and Eq.~\ref{eq:recall-probability-m}, the forgetting rate $n(t)$ and recall probability $m(t)$ is defined by the following
two coupled jump SDEs:
\begin{align*}
dn(t) =& -\alpha n(t) r(t) dN(t) + \beta n(t)(1-r(t)) dN(t) \\
dm(t) =& -n(t) m(t) dt + (1-m(t)) dN(t)
\end{align*}
with initial conditions $n(t_0) = n_0$ and $m(t_0) = m_0$.

Next, we will define an optimal cost-to-go function $J$ for the above problem, use Bellman'{}s principle of optimality to derive
the corresponding Hamilton-Jacobi-Bellman (HJB) equation~\cite{bertsekas1995dynamic}, and exploit the unique structure
of the HJB equation to find the optimal solution to the problem.
\begin{definition}
The optimal cost-to-go $J(m(t), n(t), t)$ is defined as the minimum of the expected value of the cost of going from state $(m(t), n(t))$
at time $t$ to the final state at time $t_f$.
\begin{equation}  \label{eq:cost-to-go}
  J(m(t), n(t), t) = \min_{u(t,t_f]} \EE_{(N, r)(t,t_f]} \left[ \phi(m(t_f), n(t_f)) + \int_{t}^{t_f} \ell(m(\tau), n(\tau), u(\tau))d\tau \right]
\end{equation}
\end{definition}
Now, we use Bellman'{}s principle of optimality, which the above definition allows\footnote{\scriptsize Bellman'{}s principle of optimality readily
follows using the Markov property of the recall probability $m(t)$ and forgetting rate $n(t)$.}, to break the problem into smaller subproblems,
and rewrite Eq.~\ref{eq:cost-to-go} as:
\begin{align}
J(m(t), n(t), t) &= \underset{u(t,t+dt]}{\text{min}} \EE[J(m(t+dt),n(t+dt),t+dt)] \nonumber + \ell(m(t),n(t),u(t))dt \nonumber \\
0 &= \underset{u(t,t+dt]}{\text{min}} \EE[dJ(m(t),n(t),t)] + \ell(m(t),n(t),u(t))dt, \label{eq:bellman}
\end{align}
where $dJ(m(t),n(t),t) = J(m(t+dt),n(t+dt),t+dt) - J(m(t),n(t),t)$. Then, we differentiate $J$ with respect to time $t$, $m(t)$ and $n(t)$ using
the following Lemma (proven in Appendix~\ref{app:lemma-ito}).
\begin{lemma} \label{lem:ito}
Let $x(t)$ and $y(t)$ be two jump-diffusion processes defined by the following jump SDEs:
\begin{align*}
dx(t) &= f(x(t), y(t), t)dt + g(x(t),y(t),t)z(t)dN(t)+ h(x(t),y(t),t)(1-z(t))dN(t)\\
dy(t) &= p(x(t),y(t),t)dt + q(x(t),y(t),t)dN(t)
\end{align*}
where $N(t)$ is a jump process and $z(t) \in \{0, 1\}$. If function $F(x(t), y(t), t)$ is once continuously differentiable in $x(t)$, $y(t)$
and $t$, then,
\begin{equation*}
dF(x, y, t) =(F_t + f F_x + p F_y)(x, y, t)dt + [F(x+g,y+q,t)z(t)+F(x+h,y+q,t)(1-z(t)) - F(x,y,t)]dN(t),
\end{equation*}
where for notational simplicity we dropped the arguments of the functions $f$, $g$, $h$, $p$, $q$ and argument of state variables.
\end{lemma}
Specifically, consider $x(t) = n(t)$, $y(t) = m(t)$, $z(t) = r(t)$ and $J = F$ in the above Lemma, then,
\begin{equation*}
dJ(m,n,t) = J_t(m,n,t)- nmJ_m(m,n,t)+ [J(1,(1-\alpha)n,t)r  +J(1,(1+\beta)n,t)(1-r) - J(m,n,t)]dN(t).
\end{equation*}
Then, if we substitute the above equation in Eq.~\ref{eq:bellman}, use that $\EE[dN(t)] = u(t)dt$ and $\EE[r(t)] = m(t)$,
and rearrange terms, the HJB equation follows:
\begin{equation}
0 = J_t(m,n,t)- nmJ_m(m,n,t) +  \underset{u(t,t+dt]}{\text{min}} \big\{ \ell(m,n,u) [J(1,(1-\alpha)n,t)m + J(1,(1+\beta)n,t)(1-m)-J(m,n,t)] u(t) \big\} \label{eq:hjb}
\end{equation}
\begin{algorithm}[t] 
\begin{spacing}{1.1}
\begin{algorithmic}[1]
\small
\REQUIRE Parameter $q$, $\alpha$, $\beta$, $n_0$, $t_f$
\ENSURE Next reviewing time $t$
\STATE $(u(t), n(t)) \gets (q^{-1/2}, n_0)$;
\STATE $s \gets Sample(u(t))$
\COMMENT{sample initial reviewing time}
\WHILE{$s < t_f$}
    \STATE $r(s) \gets ReviewItem(s)$ \COMMENT{\small review item, $r(s) \in \{0,1\}$}
    \STATE $n(t) \gets (1-\alpha)n(s)r(s) + (1+\beta)n(s)(1-r(s))$ \COMMENT{\small update forgetting rate}
    \STATE $m(t) \gets \exp(-(t-s) n(t))$ \COMMENT{\small update recall probability}
    \STATE $u(t) \gets q^{-1/2} (1-m(t))$ \COMMENT{\small update reviewing intensity}
    \STATE $s \gets Sample(u(t))$ \COMMENT{\small sample next reviewing time}
\ENDWHILE
\STATE \algorithmicreturn $t$ \;
\caption{The \memorize{} Algorithm}
\label{alg:memorize}
\end{algorithmic}
\end{spacing}
\end{algorithm}

To solve the above equation, we need to define the loss $\ell$.
Following the literature on stochastic optimal control~\cite{bertsekas1995dynamic}, we consider the following quadratic form,
which is nonincreasing (nondecreasing) with respect to the recall probabilities (intensities) so that it rewards learning while limiting
the number of item reviews:
\begin{equation} \label{eq:loss}
\ell(m(t),n(t),u(t)) = \frac{1}{2} (1-m(t))^2 + \frac{1}{2} q u^2(t).
\end{equation}
where $q$ is a given parameter, which trade-offs recall probability and number of item reviews.
This particular choice of loss function does not directly place a hard constraint on number of reviews---instead, it limits the number of reviews
by penalizing high reviewing intensities.
%
%
%

Under these definitions, we can find the relationship between the optimal intensity and the optimal cost by taking the derivative with respect
to $u(t)$ in Eq.~\ref{eq:hjb}:
\begin{equation*}
u^*(t) = q^{-1} \left[J(m(t),n(t),t)-J(1,(1-\alpha) n(t),t)m(t) -J(1,(1+\beta) n(t),t)(1-m(t)) \right]_{+}.
\end{equation*}
Finally, we plug in the above equation in Eq.~\ref{eq:hjb} and find that the optimal cost-to-go $J$ needs to satisfy the following nonlinear
differential equation:
\begin{align}
0 &= J_t(m(t),n(t),t)- n(t)m(t)J_m(m(t),n(t),t) + \frac{1}{2} (1-m(t))^2 \\
& - \frac{1}{2}q^{-1} \big[J(m(t),n(t),t) - J(1,(1-\alpha)n(t),t)m(t) -J(1,(1+\beta)n(t),t)(1-m(t))\big]_{\hspace{-0.8mm}+}^2. \nonumber
\end{align}
with $J(m(t_f),n(t_f), t_f) = \phi(m(t_f),n(t_f))$ as terminal condition.
To continue further, we rely on a technical Lemma (refer to Appendix~\ref{app:pro-hjb-proposals}), which derives the optimal cost-to-go $J$
for a general family of losses $\ell$.
Using this Lemma, the optimal reviewing intensity is readily given by following Theorem (proven in Appendix~\ref{app:optimal-intensity}):
\begin{theorem} \label{thm:optimal-intensity}
Given a single item, the optimal reviewing intensity for the spaced repetition problem, defined by Eq.~\ref{eq:spaced-repetition-problem-one-item},
under quadratic loss, defined by Eq.~\ref{eq:loss}, is given by $u^{*}(t)  = q^{-1/2} (1-m(t))$.
\end{theorem}
Note that the optimal intensity only depends on the recall probability, whose dynamics are given by Eqs.~\ref{eq:forgetting-rate} and~\ref{eq:recall-probability-m},
and thus allows for a very efficient procedure to sample reviewing times.
Algorithm~\ref{alg:memorize} summarizes our sampling method, which we name \memorize. Within the algorithm, $ReviewItem(s)$ returns the recall
outcome $r(s)$ of an item review at time $s$, where $r(s) = 1$ indicates the item was recalled successfully and $r(s) = 0$ indicates it was not recalled,
and $Sample(u(t))$ samples from an inhomogeneous poisson process with intensity $u(t)$ and it returns the sampled time. In practice, we sample from
an inhomogeneous poisson process using a standard thinning algorithm~\cite{lewis1979simulation}.

\xhdr{Optimizing for multiple items} Given a set of items $\Ical$ with reviewing intensities $\ub(t)$ and associated counting processes $\Nb(t)$, recall
outcomes $\rb(t)$, recall probabilities $\mb(t)$ and forgetting rates $\nbb(t)$, we can solve the spaced repetition problem defined
by Eq.~\ref{eq:spaced-repetition-problem} similarly as in the case of a single item.

More specifically, consider the following quadratic form for the loss $\ell$:
\begin{equation*}
\ell(m(t),n(t),u(t)) = \frac{1}{2} \sum_{i \in \Ical} (1-m_i(t))^2 + \frac{1}{2} \sum_{i \in \Ical} q_i u_i^2(t).
\end{equation*}
where $\{q_i\}_{i \in \Ical}$ are given parameters, which trade-off recall probability and number of item reviews and may \emph{favor} the learning of
one item over another.
Then, one can exploit the independence among items assumption to derive the optimal reviewing intensity for each item, proceeding similarly as in the
case of a single item:
\begin{theorem} \label{thm:optimal-intensity-several-items}
Given a set of items $\Ical$, the optimal reviewing intensity for each item $i \in \Ical$ in the spaced repetition problem,
defined by Eq.~\ref{eq:spaced-repetition-problem}, under quadratic loss is given by $u^{*}_i(t) = q_i^{-1/2} (1-m_i(t))$.
\end{theorem}
Finally, note that we can easily sample item reviewing times simply by running $|\Ical|$ instances of \memorize{} (Algorithm~\ref{alg:memorize}),
one per item.

\section{Experiments}
\label{sec:experiments}
\begin{figure*}[t]
  \centering
  \subfloat[Forgetting rate]{\includegraphics[width=0.25\textwidth]{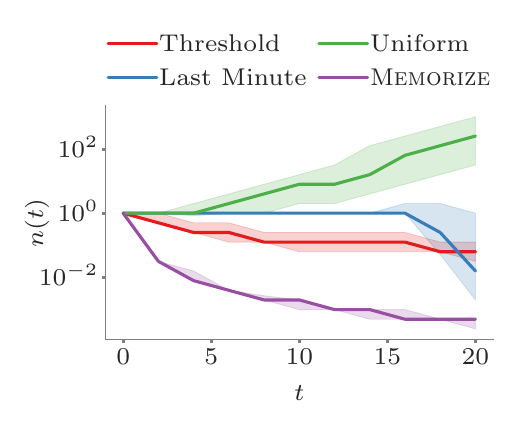}}
  \subfloat[Short-term recall prob.]{\includegraphics[width=0.25\textwidth]{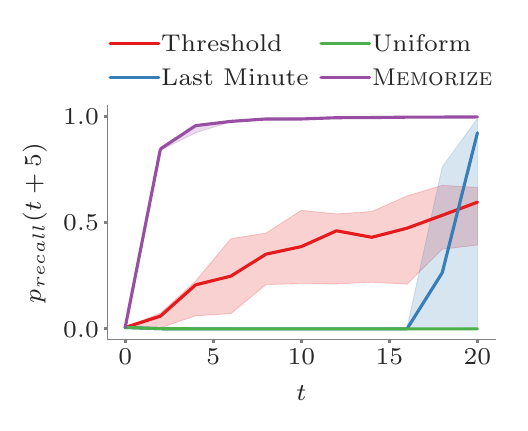}}
  \subfloat[Long-term recall prob.]{\includegraphics[width=0.25\textwidth]{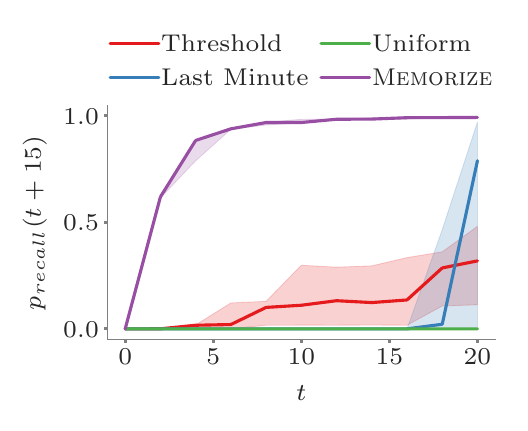}}
  \subfloat[Reviewing intensity]{\includegraphics[width=0.25\textwidth]{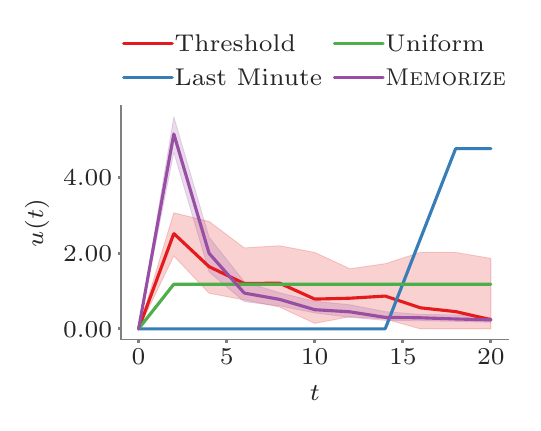}}
  \caption{Performance of \memorize{} in comparison with several baselines.
        The solid lines are median values and the shadowed regions are $30$\% confidence intervals. Short-term recall probability corresponds
        to $m(t+5)$ and long-term recall probability to $m(t+15)$.
        In all cases, we use $\alpha=0.5$, $\beta=1$, $n(0)=1$ and $t_f - t_0 = 20$. Moreover, we set $q=3 \cdot 10^{-4}$ for \memorize, $\mu = 0.6$ for the uniform
        reviewing schedule, $t_{lm} = 5$ and $\mu = 2.38$ for the last minute reviewing schedule, and $m_{th} = 0.7$ and $c = \zeta = 5$ for the threshold based
        reviewing schedule.
        Under these parameter values, the total number of reviewing events for all methods are equal (with a tolerance of $5$\%).
        }
       \label{fig:synthetic_one_card}
\end{figure*}
\subsection{Experiments on synthetic data}

In this section, our goal is analyzing the performance of \memorize{} under a controlled setting using metrics and baselines that we cannot compute in the real
data we have access to. 
%
%
%
%
%
%

\xhdr{Experimental setup}
We evaluate the performance of \memorize{} using two quality metrics: recall probability $m(t+\tau)$ at a given time in the future $t+\tau$ and forgetting rate $n(t)$.
Here, by considering high (low) values of $\tau$, we can assess long-term (short-term) retention.
Moreover, we compare the performance of our method with three baselines: (i) a \emph{uniform} reviewing schedule, which sends item(s) for review at a constant rate $\mu$;
(ii) a \emph{last minute} reviewing schedule, which only sends item(s) for review during a period $[t_{lm}, t_f]$, at a constant rate $\mu$ therein;
and (iii) a \emph{threshold} based reviewing schedule, which increases the reviewing intensity of an item by $c \exp\left((t-s)/\zeta\right)$ at time $s$, when its recall probability
reaches a threshold $m_{th}$. The threshold baseline is similar to the heuristics proposed by Metzler-Baddeley et al.~\cite{metzler2009does} and
Lindsey et al.~\cite{lindsey2014improving}.
We do not compare with the algorithm proposed by Reddy et al.~\cite{reddy2016unbounded} because, as it is specially designed for Leitner system, it assumes a discrete set of
forgetting rate values and, as a consequence, is not applicable to our (more general) setting.
Unless otherwise stated, we set the parameters of the baselines and our method such that the total number of reviewing events during $(t_0, t_f]$ are equal.

\xhdr{Solution quality}
For each method, we run $100$ independent simulations and compute the above quality metrics over time. Figure~\ref{fig:synthetic_one_card} summarizes the results,
which show that our model:
(i) consistently outperforms all the baselines in terms of both quality metrics; (ii) is more robust across runs both in terms of quality metrics and reviewing
schedule; and (iii) reduces the reviewing intensity as times goes by and the recall probability improves, as one could have expected.

\begin{figure*}[t]
  \centering
      \setlength{\tabcolsep}{0pt}
      \subfloat[Learning effort]{ \label{fig:learning-effort}
      \begin{tabular}{c c}
      \includegraphics[width=0.25\textwidth]{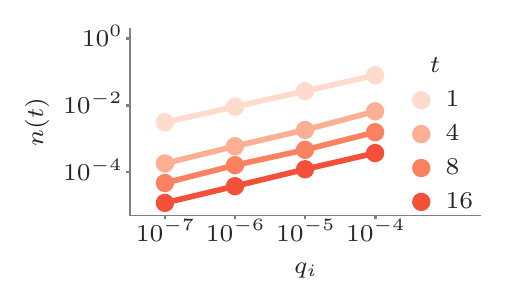} &
      \includegraphics[width=0.25\textwidth]{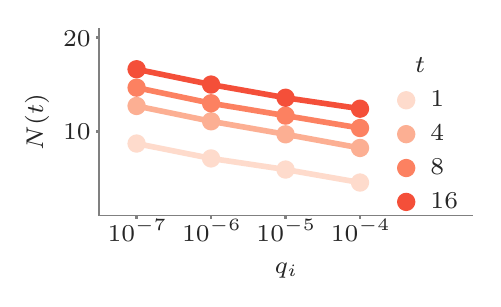} \\
      \end{tabular}
      }
      \subfloat[Item difficulty and aptitude of the learner]{ \label{fig:aptitude-learner}
      \begin{tabular}{c c}
      \includegraphics[width=0.25\textwidth]{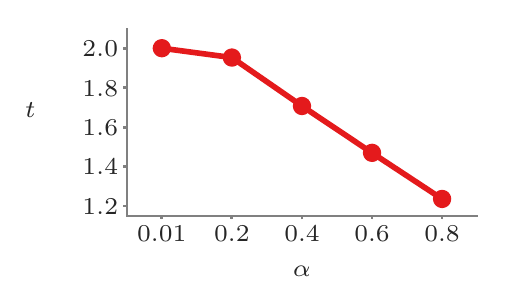}&
      \includegraphics[width=0.25\textwidth]{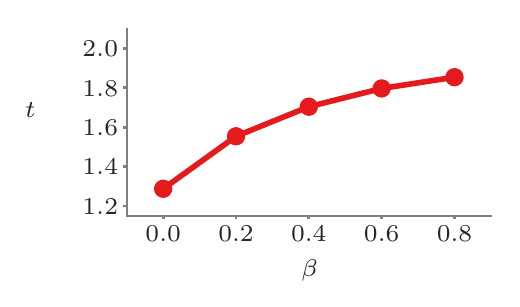}
      \end{tabular}
      }
       \caption{Learning effort, aptitude of the learner and item difficulty.
       Panel (a) shows the average forgetting rate $n(t)$ and number of reviewing events $N(t)$ for different values of the parameter $q$, which
       controls the learning effort.
       Panel (b) shows the average time the learner takes to reach a forgetting rate $n(t) = \frac{1}{2} n(0)$ for different values of the parameters
       $\alpha$ and $\beta$, which capture the aptitude of the learner and the item difficulty.
       In Panel (a), we use $\alpha=0.5$, $\beta = 1$, $n(0) = 1$ and $t_f - t_0 = 20$. In Panel (b), we use $n(0) = 20$ and $q = 0.02$.
       In both panels, error bars are too small to be seen.
       }
       \label{fig:synthetic_analysis}
       \vspace{-3mm}
\end{figure*}

\xhdr{Learning effort}
The value of the parameter $q$ controls the learning effort required by \memorize{}---the lower its value, the higher the number of reviewing events. Intuitively,
one may also expect the learning effort to influence how quickly a learner memorizes a given item---the lower its value, the quicker a learner will memorize it.
Figure~\ref{fig:learning-effort} confirms this intuition by showing the average forgetting rate $n(t)$ and number of reviewing events $N(t)$ at several times $t$
for different $q$ values.

\xhdr{Aptitude of the learner and item difficulty} The parameters $\alpha$ and $\beta$ capture the aptitude of a learner and the difficulty of the item to be learned---the
higher (lower) the value of $\alpha$ ($\beta$), the quicker a learner will memorize the item.
In Figure~\ref{fig:aptitude-learner}, we evaluate quantitatively this effect by means of the average time the learner takes to reach a forgetting rate of $n(t) = \frac{1}{2} n(0)$
using \memorize{} for different parameter values.

\subsection{Experiments on real data}\label{sec:real-experiments}
%

%
In this section, our goal is to evaluate how \emph{well} each re\-vie\-wing schedule spaces the reviews leveraging a real dataset\footnote{\scriptsize Note that it is not the objective of this
paper to evaluate the predictive power of the underlying memory models, we are relying on previous work for that~\cite{settlestrainable,wixted2007wickelgren}. However, for completeness,
we provide a series of benchmarks and evaluations for the models we used in this paper in Appendix~\ref{app:empirical-evaluation}.}.
Unlike the synthetic experiments, we cannot intervene and determine \emph{what would have happened} if a user would follow \memorize{} or any of the
baselines in the real dataset. As a consequence, measuring the performance of different algorithms is more challenging.
We overcome this difficulty by relying on likelihood comparisons to determine \emph{how closely} a (user, item) pair followed a particular re\-vie\-wing sche\-dule and
compute quality metrics that do not depend on the choice of memory model.
%
%
%
%
%
%
%
%
%
%
%

\xhdr{Datasets description} We use data gathered from Duolingo, a popular language-learning online platform\footnote{\scriptsize The dataset is available at https://github.com/duolingo/halflife-regression.}.
%
This dataset consists of $\sim$$12$ million \emph{sessions} of study, involving $\sim$5.3 million unique (user, word) pairs, which we denote by $\Dcal$, collected over the period of
two weeks.
In a single session, a user answers multiple questions, each of which contains multiple words.
Each word maps to an item $i$ and the fraction of correct recalls of sentences containing a word $i$ in the session is used as an estimate
of its recall probability at the time of the session, as in previous work~\cite{settlestrainable}. If a word is recalled perfectly during
a session then it is considered as a successful recall, \ie, $r_i(t) = 1$, and otherwise it is considered as an unsuccessful recall, \ie, $r_i(t) = 0$.
Since we can only expect the estimation of the model parameters to be accurate for users and items with enough number of reviewing events, we only consider
users with at least $30$ reviewing events and words that were reviewed at least $30$ times.
After this preprocessing step, our dataset consists of $\sim$5.2 million unique (user, word) pairs.

\xhdr{Experimental setup and methodology}
As pointed out previously, we cannot intervene in the real datasets and thus rely on likelihood comparisons to determine \emph{how closely} a (user, item) pair followed a particular reviewing
schedule. More in detail, we proceed as follows.

First, we estimate the parameters $\alpha$ and $\beta$ using half-life regression\footnote{\scriptsize Half-life $h$ is the inverse of our forgetting rate $n(t)$ multiplied by a constant.}, where
we fit a single set of parameters for all items, but a different initial forgetting rate $n_i(0)$ per item (refer to Appendix~\ref{app:empirical-evaluation} for more details).
Then, for each user, we use maximum likelihood estimation to fit the parameter $q$ in \memorize{} and the parameter $\mu$ in the uniform reviewing schedule. For the threshold based reviewing 
schedule, we fit one set of parameters for each sequence of review events using maximum likelihood estimation for the parameter $c$ and grid search for the parameter $\zeta$.

Then, we compute the likelihood of the times of the reviewing events for each (user, item) pair under the intensity given by \memorize, \ie, $u(t) = q^{-{1}/{2}} (1-m(t))$ , the intensity given by the uniform
schedule, \ie, $u(t) = \mu$, and the intensity given by the threshold based schedule, \ie, $u(t) = c \exp((t-s)/\zeta)$.
The likelihood $LL(\left\{ t_i \right\})$ of a set of reviewing events $\left\{ t_i \right\}$ given an intensity function $u(t)$ can be computed as follows~\cite{AalBorGje08}:
\begin{equation}
LL(\left\{ t_i \right\}) = \sum_{i} \log{u(t_i)} - \int_{0}^{T} u(t)\,dt.
\nonumber
\end{equation}
This allows us to determine how closely a (user, item) pair follows a particular reviewing schedule\footnote{\scriptsize Duolingo uses hand-tuned spaced repetition algorithms, which propose re\-vie\-wing
times to the users. 
However, since users often do not perform reviews exactly at the recommended times, some pairs will be closer to uniform than threshold or \memorize{} and viceversa.}, as shown in Figure~\ref{fig:sample_trace}. Distribution of the likelihood values under each reviewing schedule is provided in Appendix~\ref{app:empirical-likelihood}.
We do not compare to the last minute baseline since in Duolingo there is no terminal time $t_f$ which users target. Additionally, in many (user, item) pairs, the first review takes place close to $t=0$ and thus the last minute baseline is equivalent to the uniform reviewing schedule.
   \begin{figure}[t]
     \centering
       \begin{tabular}{lcc}
         \memorize&\includegraphics[width=0.31\textwidth,valign=c]{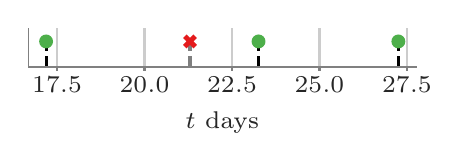}\\
         Threshold&\includegraphics[width=0.31\textwidth,valign=c]{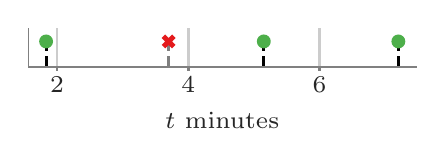}\\
         Uniform&\includegraphics[width=0.31\textwidth,valign=c]{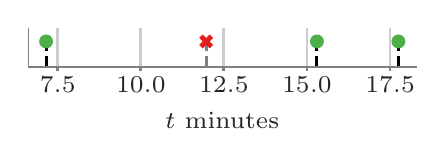}\\
       \end{tabular}
    \caption{
    Examples of (user, item) pairs whose corresponding reviewing times have high likelihood under \memorize{} (top), uniform reviewing schedule (bottom) and threshold based reviewing
    schedule (middle).
   %
   %
   In every figure, each candle stick corresponds to a reviewing event with a green circle (red cross) if the recall was (un)successful, and time $t=0$ corresponds to the first
   time the user is exposed to the item in our dataset, which may or may not correspond with the first reviewing event.
   The pairs whose reviewing times follow more closely \memorize{} or the threshold based schedule tend to increase the time interval between reviews every time a recall is successful while, in
   contrast, the uniform reviewing schedule does not. \memorize{} tends to space the reviews more than the threshold based schedule, achieving the same recall pattern with less effort.
   }
   \label{fig:sample_trace}
   \vspace{-2mm}
 \end{figure}

Finally, since measurements of the future recall probability $m(t + \tau)$ are not forthcoming and depend on the memory model of choice, we concentrate on the following alternative
quality me\-trics, which do not depend on the particular choice of memory model:
\begin{itemize}[leftmargin=6mm]
\item[(a)] \emph{Effort}: for each (user, item), we measure the effort by means of the empirical estimate of the inverse of the total reviewing period, \ie, $\hat{e} = 1 / (t_n - t_1)$. The lower
the effort, the less burden on the user, allowing her to learn more items simultaneously.

\item[(b)] \emph{Empirical forgetting rate}: for each (user, item), we compute an empirical estimate of the forgetting rate by the time $t_n$ of the last reviewing event, \ie,
$\hat{n} = -\log(\hat{m}(t_n))/(t_n - t_{n-1})$.
Here, note that the estimate of the forgetting rate only depends on the observed data (not model/methods parameters).
For a more fair comparison across items, we normalize each empirical forgetting rate using the average empirical initial forgetting rate of the corresponding item at the beginning of the observation window, \ie, for an
item $i$, $\hat{n}_0 = |\left\{ u : (u, i) \in \Dcal \right\}|^{-1} \sum_{u : (u, i) \in \Dcal} \hat{n}_{0,u}$ where $\hat{n}_{0,u} = -\log(\hat{m}(t_{u,1}))/\allowbreak(t_{u,1} -t_{u,0})$.
\end{itemize}
Given a particular recall pattern, the lower the above quality metrics, the more effective the reviewing schedule.
\begin{figure}[t]
   \centering
   \captionsetup[subfigure]{labelformat=empty}
   \subfloat[][]{
   \setlength{\tabcolsep}{2.9mm}
   \begin{tabular}{l|c|c|c@{\hskip -1mm}|c|c}
   &\multicolumn{2}{c|}{\textbf{Effort, $\hat{e}$}}&&\multicolumn{2}{|c}{\textbf{Emp. forgetting, $\hat{n}$}}\\
   \hhline{======}
      \textbf{Pattern} & $\frac{\hat{e}_M}{\hat{e}_T}$ & $\frac{\hat{e}_M}{\hat{e}_U}$  &&$\frac{\hat{n}_M}{\hat{n}_T}$ & $\frac{\hat{n}_M}{\hat{n}_U}$ \\[0.2ex]
      \hline
     \recall\recall\recall                & $0.00^{*}$             & $0.01^{*}$ &&$0.11^{*}$ & $0.11^{*}$ \\
    \recall\recall\recall\recall         & $0.01^{*}$             & $0.04^{*}$ &&$0.14^{*}$ & $0.15^{*}$ \\
     \recall\recall\recall\recall\recall  & $0.02^{*}$             & $0.07^{*}$ &&$0.21^{*}$ & $0.22^{*}$  \\
      \forget\recall\recall                & $0.00^{*}$             & $0.03^{*}$  &&$0.10^{*}$ & $0.10^{*}$ \\
      \forget\recall\recall\recall         & $0.03^{*}$             & $0.06^{*}$ &&$0.13^{*}$ & $0.14^{*}$ \\
      \forget\forget\recall\recall         & $0.03^{*}$             & $0.05^{*}$ &&$0.11^{*}$ & $0.12^{*}$  \\
     \forget\forget\forget\recall\recall  & $0.03^{*}$   & $0.05^{*}$ &&$0.12^{*}$ & $0.12^{*}$\\
   \end{tabular}%
}
   \caption{Performance of \memorize{} (M) in comparison with the uniform (U) and threshold (T) based reviewing schedules. Each row corresponds to a different recall pattern, depicted in the
 first column, where markers denote ordered recalls, green circles indicate successful recall and red crosses indicate unsuccessful ones. In the second column, each cell value corresponds to the ratio between
 the average effort for the top 25\% pairs in terms of likelihood for the uniform (or threshold based) schedule and \memorize. In the right column, each cell value corresponds to the ratio between the median
 empirical forgetting rate for the same pairs for the uniform (or threshold based) schedule and \memorize.
 In both metrics, if the ratio is smaller than $1$, \memorize{} is more effective than the uniform (or threshold based) schedule for the corresponding pattern.
 The symbol $*$ indicates whether the change is significant with p-value $<0.01$ using the Kolmogrov-Smirnov 2-sample test.
 }
 \label{tbl:real_data}
 \vspace{-1mm}
\end{figure}

\xhdr{Results}
We first group (user, item) pairs by their recall pattern, \ie, the sequence of successful ($r = 1$) and unsuccessful ($r = 0$) recalls over time---if two pairs have the
same recall pattern, then they have the same number of reviews and changes in their forgetting rates $n(t)$.
For each recall pattern in our observation window, we pick the top 25\% pairs in terms of likelihood for each method and compute the average effort and
empirical forgetting rate, as defined above.
Figure~\ref{tbl:real_data} summarizes the results for the most common recall patterns\footnote{\scriptsize Results are qualitatively similar for other recall patterns.}, where we
report the ratio between the effort and empirical forgetting rate values achieved by the uniform and threshold based reviewing schedules and the values achieved by \memorize. That means, if the reported
value is smaller than $1$, \memorize{} is more effective for the corresponding pattern.
We find that both in terms of effort and empirical forgetting rate, \memorize{} outperforms the uniform and threshold based reviewing schedules for all recall
patterns.
For example, for the recall pattern consisting of two unsuccessful recalls followed by two successful recalls (red-red-green-green), \memorize{} achieves $0.05$$\times$
lower effort and $0.12$$\times$ lower empirical forgetting rate than the second competitor.

\begin{figure}[t]
  \centering
  \begin{tabular}{c}
  \includegraphics[width=0.49\textwidth]{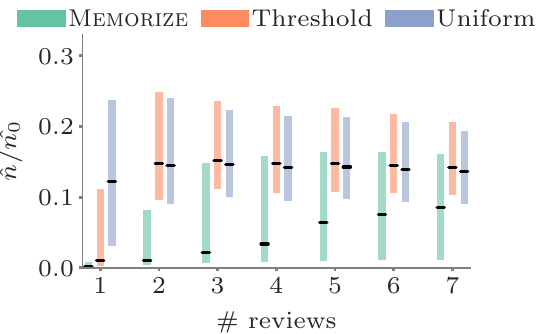}
  \end{tabular}
  \caption{Average empirical forgetting rate for the top 25\% pairs in terms of likelihood for \memorize{}, the uniform reviewing schedule and the threshold based reviewing schedule for sequences with different number
  of reviews. Boxes indicate 25\% and 75\% quantiles and solid lines indicate median values, where lower values indicate better performance. In all cases, the competitive advantage \memorize{} achieves
  is statistically significant (p-value $<0.01$ using the Kolmogrov-Smirnov 2-sample test).
  }\label{fig:real_data_review}
\end{figure}

Next, we group (user, item) pairs by the number of reviews during a fixed period of time, \ie, we control for the effort,  pick the top 25\% pairs in terms of likelihood for each method and
compute the average empirical forgetting rate.
Figure~\ref{fig:real_data_review} summarizes the results for sequences with up to seven given reviews since the beginning of the observation window, where lower values indicate
better performance. The results show that \memorize{} offers a competitive advantage with respect to the other baselines, which is statistically significant.

\section{Conclusions}
\label{sec:conclusions}
In this paper, we have first introduced a novel representation of spaced repetition using the framework of marked temporal point processes and
SDEs with jumps and then designed a framework that exploits this novel representation to cast the design of spaced repetition algorithms as a stochastic
optimal control problem of such SDEs.
For ease of exposition, we have considered only two memory models, exponential and power-law forgetting curves, and a quadratic loss function, however, 
our framework is agnostic to this particular modeling choices and it provides a set of novel techniques to find reviewing schedules that are optimal \emph{under} 
a given choice of memory model and loss.
We experimented on both synthetic and real data gathered from Duolingo, a popular language-learning online platform, and showed that our framework 
may be able to help learners memorize more effectively than alternatives.

There are many interesting directions for future work. For example,
it would be interesting to perform large scale interventional experiments to assess the performance of our algorithm in comparison with existing
spaced repetition algorithms deployed by, \eg, Duolingo.
Moreover, in our work, we consider a particular quadratic loss, however, it would be useful to derive optimal reviewing intensities for other (non-quadratic) losses
cap\-tu\-ring particular learning goals.
We assumed that, by reviewing an item, one can only influence its recall probability and forgetting rate. However, items may be dependent
and thus, by reviewing an item, one can influence the recall probabilities and forgetting rates of several items.
Finally, it would be very interesting to allow for reviewing events to be composed of groups of items, some reviewing times to be preferable over others,
and then derive both the optimal reviewing schedule and optimal grouping of items.

\bibliographystyle{abbrv}
\bibliography{refs}

\appendix
\section{Proof of Proposition~\ref{pro:recall-probability}} \label{app:pro-recall-probability}

According to Eq.~\ref{eq:recall-probability}, the recall probability $m(t)$ depends on the forgetting
rate, $n(t)$, and the time elapsed since the last review, $D(t) := t - t_r$. Moreover, we can readily
write the differential of $D(t)$ as $dD(t) = dt - D(t) dN(t)$.

We define the vector $\Xb(t) = [n(t), D(t)]^{T}$. Then, we use Eq.~\ref{eq:forgetting-rate} and
It\"{o}'{}s calculus~\cite{hanson2007} to compute its differential:
\begin{align*}
d\Xb(t) &\stackequal{\text{dt}}{zol} \fb(\Xb(t),t)dt+\hb(\Xb(t),t)dN(t) \\
\fb(\Xb(t), t) &= \icol{0 \\ 1} \\
\hb(\Xb(t), t) &= \icol{-\alpha n(t) r(t) + \beta n(t)(1-r(t)) \\ -D(t)}
\end{align*}

Finally, using again It\"{o}'{}s calculus and the above differential, we can compute the differential
of the recall probability $m(t) = e^{-n(t)D(t)} := F(\Xb(t))$ as follows:
\begin{align*}
  dF(\Xb(t)) &= F(\Xb(t+dt))-F(\Xb(t)) \\
           &= F(\Xb(t)+d\Xb(t))-F(\Xb(t)) \\
           &\stackequal{\text{dt}}{zol} (f^T F_{\Xb}(\Xb(t)))dt + F(\Xb(t) + \hb(\Xb(t), t)dN(t))-F(\Xb(t)) \\
           &\stackequal{\text{dt}}{zol} (f^T F_{\Xb}(\Xb(t)))dt + (F(\Xb(t)+h(\Xb(t), t))-F(\Xb(t)))dN(t) \\
           &= (e^{-(D(t)-D(t))n(t)(1 + \alpha r_i(t) - \beta (1-r_i(t)))}-e^{-D(t)n(t)})dN(t) -n(t)e^{-D(t)n(t)}dt\\
           &= -n(t)e^{-D(t)n(t)}dt + (1-e^{-D(t)n(t)})dN(t) \\
           &= -n(t)F(X(t))dt + (1-F(X(t)))dN(t) \\
           &= -n(t)m(t)dt + (1-m(t))dN(t).
\end{align*}

\section{Proof of Lemma~\ref{lem:ito}} \label{app:lemma-ito}

According to the definition of differential,
\begin{align*}
dF := dF(x(t), y(t), t) &= F(x(t+dt), y(t+dt), t+dt) - F(x(t), y(t), t) \\
&=F(x(t)+dx(t), y(t)+dy(t), t+dt) - F(x(t), y(t), t).
\end{align*}
Then, using It\"{o}'{}s calculus, we can write
\begin{align} \label{eq:lem-3-eq}
& dF \stackequal{\text{dt}}{zol} F(x+ fdt + g, y + pdt + q, t + dt)dN(t)z +F(x + fdt + h, y + pdt + q, t + dt)dN(t)(1-z) \nonumber \\
&+F(x + fdt, y + pdt, t + dt)(1-dN(t)) - F(x,y,t)
\end{align}
where for notational simplicity we drop arguments of all functions except $F$ and
$dN$. Then, we expand the first three terms:
\begin{align*}
F(x + fdt + g, y + pdt + q, t + dt) &= F(x + g, y + q, t) + F_x(x + g, y + q, t) fdt + F_y(x + g, y + q, t) pdt \\
&+ F_t(x + g, y + q, t) dt\\
F(x + fdt + h, y + pdt + q, t + dt) &= F(x + h, y + q, t) + F_x(x + h,y + q,t) fdt  +F_y(x + h, y + q, t) pdt \\
&+ F_t(x + h, y + q, t) dt\\
F(x + fdt, y + pdt, t + dt) &= F(x, y, t) + F_x(x, y, t) fdt + F_y(x, y, t) pdt + F_t(x, y, t) dt
\end{align*}
using that the bilinear differential form $dt\,dN(t) = 0$. Finally, by substituting the above three equations into
Eq.~\ref{eq:lem-3-eq}, we conclude that:
\begin{align*}
dF(x(t), y(t), t) &= (F_t + f F_x + p F_y)(x(t), y(t), t)dt +\big[F(x+g,y+q,t)z(t) +F(x+h,y+q,t)(1-z(t))\\
&\quad -F(x,y,t)\big]dN(t),
\end{align*}

\section{Lemma~\ref{lem:hjb-proposals}} \label{app:pro-hjb-proposals}

\begin{lemma}  \label{lem:hjb-proposals}
Consider the following family of losses with parameter $d > 0$,
\begin{align}
\ell_d(m(t),n(t),u(t)) &~= h_d(m(t),n(t)) + g_d^2(m(t),n(t)) + \frac{1}{2} q u(t)^2, \nonumber \\
g_d(m(t),n(t)) &~= 2^{-1/2} \left[c_2 \frac{\log(d)}{-m(t)^2 + 2m(t) - d} - c_2\frac{\log(d)}{1-d} + c_1 m(t) \log\left( \frac{1+\beta}{1-\alpha}\right) - c_1\log(1+\beta)\right]_{\hspace{-0.8mm}+}, \nonumber \\
h_d(m(t),n(t)) &~= {-\sqrt{q}} m(t) n(t) c_2 \frac{(-2m(t) + 2) \log(d)}{(-m(t)^2+2m(t)-d)^2}. \label{eq:family}
\end{align}
where $c_1, c_2 \in \RR$ are arbitrary constants. Then, the cost-to-go $J_d(m(t),n(t),t)$ that satisfies
the HJB equation, defined by Eq.~\ref{eq:hjb}, is given by:
\begin{equation} \label{eq:family-cost-to-go}
J_d(m(t),n(t),t) = \sqrt{q} \left(c_1 \log(n(t)) + c_2 \frac{\log(d)}{-m(t)^2 + 2m(t) - d}\right)
\end{equation}
and the optimal intensity is given by:
\begin{equation*}
u_d^*(t) = q^{-1/2} [c_2 \frac{\log(d)}{-m(t)^2 + 2m(t) - d}- c_2\frac{\log(d)}{1-d} + c_1 m(t) \log\left( \frac{1+\beta}{1-\alpha}\right) - c_1\log(1+\beta)]_{\hspace{-0.8mm}+}.
\end{equation*}

\begin{proof}
Consider the family of losses defined by Eq.~\ref{eq:family} and the functional form for the cost-to-go defined by Eq.~\ref{eq:family-cost-to-go}.
Then, for any parameter value $d > 0$, the optimal intensity $u^{*}_d(t)$ is given by
\begin{align*}
u_d^*(t) &= q^{-1} \left[J_d(m(t),n(t),t)-J_d(1,(1-\alpha)n(t),t)m(t) -J_d(1,(1+\beta) n(t),t)(1-m(t)) \right]_{\hspace{-0.8mm}+} \\
&= q^{-1/2} \left[ c_2 \frac{\log(d)}{-m^2 + 2m - d}  - c_2\frac{\log(d)}{1-d} +c_1 m(t) \log\left( \frac{1+\beta}{1-\alpha}\right) - c_1\log(1+\beta)\right]_{\hspace{-0.8mm}+},
\end{align*}
and the HJB equation is satisfied:
\begin{align*}
&\frac{\partial J_d(m,n,t)}{\partial t} - mn \frac{\partial J_d(m,n,t)}{\partial m} + h_d(m,n) + g_d^2(m,n)- \frac{1}{2}q^{-1}\big(J_d(m,n,t)\\
&-J_d(1,(1-\alpha)n,t)m - J_d(1,(1+\beta)n,t)(1-m)\big)_{\hspace{-0.8mm}+}^2\\
&= \sqrt{q} mn c_2 \frac{(-2m + 2) \log(d)}{(-m^2+2m-d)^2} + h_d(m,n) + g_d^2(m,n) - \frac{1}{2}\left[ c_1 \log( n ) + c_2 \frac{\log(d)}{-m^2 + 2m - d} \right. \\
& \quad \left. - m \left(c_1 \log(n(1-\alpha)) + c_2 \frac{\log(d)}{1 - d} \right) - (1-m) \left(c_1 \log(n(1+\beta))  + c_2 \frac{\log(d)}{1 - d}\right) \right]_{\hspace{-0.8mm}+}^2 \\
&= \sqrt{q} mn c_2 \frac{(-2m + 2) \log(d)}{(-m^2+2m-d)^2} \underbrace{-\sqrt{q}  mn c_2 \frac{(-2m + 2) \log(d)}{(-m^2+2m-d)^2}}_{h_d(m,n)} \\
& \quad -\frac{1}{2}\big[ c_2 \frac{\log(d)}{-m^2 + 2m - d} - c_2 \frac{\log(d)}{1 - d} + c_1m \log(\frac{1+\beta}{1-\alpha}) - c_1 \log(1+\beta)\big]_{\hspace{-0.8mm}+}^2\\
%
& \quad +\frac{1}{2}\big[ c_2 \frac{\log(d)}{-m^2 + 2m - d}  - c_2 \frac{\log(d)}{1 - d} + c_1m \log(\frac{1+\beta}{1-\alpha}) - c_1 \log(1+\beta)\big]_{\hspace{-0.8mm}+}^2
\end{align*}
where for notational simplicity $m = m(t)$, $n = n(t)$ and $u = u(t)$.
\end{proof}
\end{lemma}



\section{Proof of Theorem~\ref{thm:optimal-intensity}} \label{app:optimal-intensity}
Consider the family of losses defined by Eq.~\ref{eq:family} in Lemma~\ref{lem:hjb-proposals} whose optimal intensity
is given by:
\begin{align*}
u_d^*(t) &= q^{-1/2} \left[ c_2 \frac{\log(d)}{-m^2 + 2m - d}  - c_2\frac{\log(d)}{1-d} + c_1 m(t) \log\left( \frac{1+\beta}{1-\alpha}\right) - c_1\log(1+\beta)\right]_{\hspace{-0.8mm}+}.
\end{align*}
Now, set the constants $c_1, c_2 \in \RR$ to the following values:
\begin{equation*}
   c_1 = \frac{-1}{\log\left(\frac{1+\beta}{1-\alpha}\right)} \quad
   c_2 = \frac{-\log\left( 1 - \alpha \right)}{\log\left(\frac{1+\beta}{1-\alpha}\right)}.
\end{equation*}
Since the HJB equation is satisfied for any value of $d > 0$, we can recover the quadratic loss $l(m,n,u)$ and derive its
corresponding optimal intensity $u^{*}(t)$ using point wise convergence:
\begin{align*}
l(m(t), n(t), u(t)) & = \lim_{d \rightarrow 1} l_d(m(t), n(t), u(t)) = \frac{1}{2} (1-m(t))^2 + \frac{1}{2} q u^2(t), \\
u^{*}(t) &= \lim_{d \rightarrow 1} u_d^{*}(t) = q^{-1/2} (1-m(t)),
\end{align*}
where we used that $\lim_{d \rightarrow 1} \frac{\log(d)}{1-d} = -1$ (L'{}Hospital'{}s rule).

%
%
%
%
%
%

This concludes the proof.

\section{The \memorize{} Algorithm under the Power-Law Forgetting Curve Model} \label{app:power-law-derivations}

In this section, we first derive the optimal reviewing schedule under the power-law forgetting curve model and then validate such schedule using the same Duolingo dataset as in the main
section of the paper.

\xhdr{Problem formulation and algorithm}
Under the power-law forgetting curve model, the probability of recalling an item $i$ at time $t$ is given by~\cite{wixted2007wickelgren}:
\begin{equation} \label{eq:recall-probability-pl-app}
m_i(t) := \PP(r_i(t)) = (1 + \omega (t - t_r) )^{-n_i(t)},
\end{equation}
where $t_r$ is the time of the last review, $n_i(t) \in RR^{+}$ is the forgetting rate and $\omega$ is a time scale parameter.
%

Similarly as in Proposition~\ref{pro:recall-probability} for the exponential forgetting curve model, we can express the dynamics of the recall probability $m_i(t)$ by
means of a SDE with jumps:
\begin{equation} \label{eq:recall-probability-dm-pl}
dm_i(t) = -\frac{n_i(t)m_i(t) \omega dt}{(1+ \omega D_i(t))} + (1-m_i(t)) dN_i(t)
\end{equation}
where $D_i(t) := t- t_r$ and thus the differential of $D_i(t)$ is readily given by $d D_i(t) = dt - D_i(t) dN_i(t)$.

Next, similarly as in the case of the exponential forgetting curve model in the main paper, we consider a single item with $n_i(t) = n(t)$, $m_i(t) = m(t)$, $D_i(t) = D(t)$ and $r_i(t) = r(t)$, and
adapt Lemma~\ref{lem:ito} to the power-law forgetting curve model as follows:
\begin{lemma} \label{lem:ito-pl}
Let $x(t)$ and $y(t)$, $k(t)$ be three jump-diffusion processes defined by the following jump SDEs:
\begin{align*}
dx(t) =& f(x(t), y(t), t)dt + g(x(t),y(t),t)z(t)dN(t) + h(x(t),y(t),t)(1-z(t))dN(t)\\
dy(t) =& p(x(t),y(t),t)dt + q(x(t),y(t),t)dN(t)\\
dk(t) =& s(x(t),y(t),k(t),t)dt+v(x(t),y(t),k(t),t)dN(t)
\end{align*}
where $N(t)$ is a jump process and $z(t) \in \{0, 1\}$. If function $F(x(t), y(t), k(t), t)$ is once continuously differentiable in $x(t)$, $y(t)$, $z(t)$
and $t$, then,
\begin{align*}
dF(x, y, k, t) & =(F_t + f F_x + p F_y + sF_s)(x, y, k, t)dt + [F(x+g,y+q,k+v,t)z(t) \\&+F(x+h,y+q,k+v,t)(1-z(t))-F(x,y,t)]dN(t),
\end{align*}
where for notational simplicity we dropped the arguments of the functions $f$, $g$, $h$, $p$, $q$, $s$, $v$ and argument of state variables.
\end{lemma}

Then, if we consider $x(t) = n(t)$, $y(t) = m(t)$, $k(t)=D(t)$, $z(t) = r(t)$ and $J = F$ in the above Lemma, the differential
of the optimal cost-to-go is readily given by
\begin{align*}
dJ(m,n,t) &= J_t(m,n,t)- \frac{\omega nm}{1+\omega D}J_m(m,n,D,t) + J_{D}(m,n,D,t)+ [J(1,(1-\alpha)n,0,t)r(t) \\
&\quad +J(1,(1+\omega)n,0,t)(1-r)-J(m,n,D,t)]dN(t).
\end{align*}

Moreover, under the same loss function $\ell(m(t),n(t),u(t))$ as in Eq.~\ref{eq:loss}, it is easy to show that the optimal cost-to-go $J$ needs to satisfy the following
nonlinear partial differential equation:
\begin{align}
0 &= J_t(m(t),n(t),t)- \frac{\omega n(t)m(t)}{1+\omega D}J_m(m(t),n(t),t) + \quad J_{D}(m(t),n(t),t) +  \frac{1}{2} (1-m(t))^2 \nonumber\\
&\quad -\frac{1}{2}q^{-1}\big(J(m(t),n(t),t) - J(1,(1-\alpha)n(t),t)m(t) - J(1,(1+\beta)n(t),t)(1-m(t))\big)_{\hspace{-0.8mm}+}^2. \label{eq:hjb-pl}
\end{align}

Then, we can adapt Lemma~\ref{lem:hjb-proposals} to derive the optimal sche\-du\-ling policy for a single item under the power-law forgetting
curve model:

\begin{lemma}  \label{lem:hjb-proposals-pl}
Consider the following family of losses with parameter $d > 0$,
\begin{align}
\ell_d(m(t),n(t),D(t),u(t)) &~= h_d(m(t),n(t),D(t)) + g_d^2(m(t),n(t)) + \frac{1}{2} q u(t)^2, \nonumber \\
g_d(m(t),n(t)) &~= 2^{-1/2} \left[c_2 \frac{\log(d)}{-m(t)^2 + 2m(t) - d} - c_2\frac{\log(d)}{1-d} + c_1 m(t) \log\left( \frac{1+\beta}{1-\alpha}\right) - c_1\log(1+\beta)\right]_{\hspace{-0.8mm}+}, \nonumber \\
h_d(m(t),n(t)) &~= {-\sqrt{q}} \frac{\omega n(t) m(t)}{1+\omega D(t)} c_2 \frac{(-2m(t) + 2) \log(d)}{(-m(t)^2+2m(t)-d)^2}. \label{eq:family-pl}
\end{align}
where $c_1, c_2 \in \RR$ are arbitrary constants. Then, the cost-to-go $J_d(m(t),n(t),t)$ that satisfies
the HJB equation, defined by Eq.~\ref{eq:hjb-pl}, is given by:
\begin{align} \label{eq:family-cost-to-go-pl}
J_d(m(t),n(t),D(t),t) = \sqrt{q} \left(c_1 \log(n(t)) + c_2 \frac{\log(d)}{-m(t)^2 + 2m(t) - d}\right)
\end{align}
which is independent of $D(t)$, and the optimal intensity is given by:
\begin{align*}
u_d^*(t) = q^{-1/2} [c_2 \frac{\log(d)}{-m(t)^2 + 2m(t) - d}- c_2\frac{\log(d)}{1-d} + c_1 m(t) \log\left( \frac{1+\beta}{1-\alpha}\right) - c_1\log(1+\beta)]_{\hspace{-0.8mm}+}.
\end{align*}

\begin{proof}
Consider the family of losses defined by Eq.~\ref{eq:family-pl} and the functional form for the cost-to-go defined by Eq.~\ref{eq:family-cost-to-go-pl}.
Then, for any parameter value $d > 0$, the optimal intensity $u^{*}_d(t)$ is given by
\begin{align*}
u_d^*(t) &= q^{-1} \left[J_d(m(t),n(t),t)-J_d(1,(1-\alpha)n(t),t)m(t) -J_d(1,(1+\beta) n(t),t)(1-m(t)) \right]_{\hspace{-0.8mm}+} \\
&= q^{-1/2} \left[ c_2 \frac{\log(d)}{-m^2 + 2m - d}  - c_2\frac{\log(d)}{1-d}  +c_1 m(t) \log\left( \frac{1+\beta}{1-\alpha}\right) - c_1\log(1+\beta)\right]_{\hspace{-0.8mm}+},
\end{align*}
and the HJB equation is satisfied:
\begin{align*}
&\frac{\partial J_d(m,n,t)}{\partial t} - \frac{\omega n m}{1+\omega D}  \frac{\partial J_d(m,n,t)}{\partial m} + h_d(m,n) + g_d^2(m,n)- \frac{1}{2}q^{-1}\big(J_d(m,n,t)\\
&-J_d(1,(1-\alpha)n,t)m - J_d(1,(1+\beta)n,t)(1-m)\big)_{\hspace{-0.8mm}+}^2\\
&= \sqrt{q} \frac{\beta c_2 n m  }{1+\omega D}  \frac{(-2m + 2) \log(d)}{(-m^2+2m-d)^2} + h_d(m,n) + g_d^2(m,n)  - \frac{1}{2}\left[ c_1 \log( n ) + c_2 \frac{\log(d)}{-m^2 + 2m - d} \right. \\
& \quad \left. - m \left(c_1 \log(n(1-\alpha)) + c_2 \frac{\log(d)}{1 - d} \right) - (1-m) \left(c_1 \log(n(1+\beta))  + c_2 \frac{\log(d)}{1 - d}\right) \right]_{\hspace{-0.8mm}+}^2 \\
&= \sqrt{q} \frac{\omega n m c_2 }{1+\omega D(t)} \frac{(-2m + 2) \log(d)}{(-m^2+2m-d)^2} \underbrace{-\sqrt{q}  \frac{\omega n m c_2 }{1+\omega D}  \frac{(-2m + 2) \log(d)}{(-m^2+2m-d)^2}}_{h_d(m,n)} \\
& \quad -\frac{1}{2}\big[ c_2 \frac{\log(d)}{-m^2 + 2m - d} - c_2 \frac{\log(d)}{1 - d} + c_1m \log(\frac{1+\beta}{1-\alpha}) - c_1 \log(1+\beta)\big]_{\hspace{-0.8mm}+}^2\\
%
& \quad +\frac{1}{2}\big[ c_2 \frac{\log(d)}{-m^2 + 2m - d}  - c_2 \frac{\log(d)}{1 - d} + c_1m \log(\frac{1+\beta}{1-\alpha}) - c_1 \log(1+\beta)\big]_{\hspace{-0.8mm}+}^2
\end{align*}
where for notational simplicity $m = m(t)$, $n = n(t)$, $D = D(t)$ and $u = u(t)$.
\end{proof}
\end{lemma}
\begin{figure}[t]
   \centering
   \captionsetup[subfigure]{labelformat=empty}
\subfloat[][]{
\setlength{\tabcolsep}{2.9mm}
\begin{tabular}{l|c|c|c@{\hskip -1mm}|c|c}
&\multicolumn{2}{c|}{\textbf{Effort, $\hat{e}$}}&&\multicolumn{2}{|c}{\textbf{Emp. forgetting,  $\hat{n}$}}\\
\hhline{======}
   \textbf{Pattern} & $\frac{\hat{e}_M}{\hat{e}_T}$ & $\frac{\hat{e}_M}{\hat{e}_U}$  &&$\frac{\hat{n}_M}{\hat{n}_T}$ & $\frac{\hat{n}_M}{\hat{n}_U}$ \\[0.2ex]
   \hline
  \recall\recall\recall                & $0.00^{*}$             & $0.02^{*}$ &&$0.11^{*}$ & $0.11^{*}$ \\
 \recall\recall\recall\recall         & $0.01^{*}$             & $0.05^{*}$ &&$0.14^{*}$ & $0.15^{*}$ \\
  \recall\recall\recall\recall\recall  & $0.02^{*}$             & $0.07^{*}$ &&$0.20^{*}$ & $0.20^{*}$  \\
   \forget\recall\recall                & $0.01^{*}$             & $0.03^{*}$  &&$0.10^{*}$ & $0.10^{*}$ \\
   \forget\recall\recall\recall         & $0.04^{*}$             & $0.07^{*}$ &&$0.13^{*}$ & $0.13^{*}$ \\
   \forget\forget\recall\recall         & $0.03^{*}$             & $0.06^{*}$ &&$0.11^{*}$ & $0.11^{*}$  \\
  \forget\forget\forget\recall\recall  & $0.04^{*}$   & $0.06^{*}$ &&$0.13{*}$ & $0.13^{*}$\\
\end{tabular}%
}
\vspace{-3mm}
\caption{Performance of \memorize{} (M) in comparison with uniform (U) and threshold (T) based reviewing schedule for the power-law forgetting curve model. Each row of each table corresponds to a different recall
pattern, depicted in the first column, where markers denote ordered recalls, green circles indicate successful recall and red crosses indicate unsuccessful ones. In the second column, each cell value corresponds to
the ratio between the average effort for the top 25\% pairs in terms of likelihood for the uniform (or threshold based) schedule and \memorize. In the right column, each cell value corresponds to the ratio between the
median empirical forgetting rate.
%
%
%
The symbol $*$ indicates whether the change is significant with p-value $<0.01$ using the Kolmogrov-Smirnov 2-sample test.
}
\label{tbl:real_data-pl}
\label{-3mm}
\end{figure}
\begin{figure}[t]
  \centering
  \begin{tabular}{c}
  \includegraphics[width=0.46\textwidth]{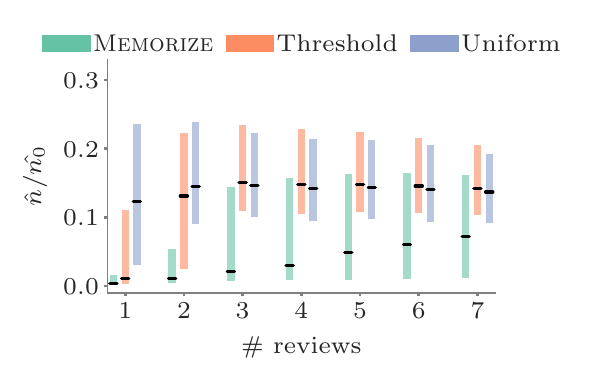}
  \end{tabular}
        \caption{Performance of \memorize{} with the power-law forgetting curve model compared to the uniform and threshold schedules. Boxes indicate 25\% and 75\% quantiles and
        solid lines indicate median value. Lower values indicate better performance. In all cases, the competitive advantage \memorize{} achieves
        is statistically significant (p-value $<0.01$ using the Kolmogrov-Smirnov 2-sample test).
        }
       \label{fig:real_data_review-pl}
       \vspace{-3mm}
\end{figure}

Finally, reusing Theorem~\ref{thm:optimal-intensity}, the optimal reviewing intensity for a single item under the power-law forgetting curve
model is given by
\begin{align*}
u^{*}(t) &= \lim_{d \rightarrow 1} u_d^{*}(t) = q^{-1/2} (1-m(t)).
\end{align*}
It is then straightforward to derive the optimal reviewing intensity for a set of items, which adopts the same form as in Theorem~\ref{thm:optimal-intensity-several-items}.

\xhdr{Experimental evaluation} In this section, the dataset, baselines, experimental setup and quality metrics are the same as in the main paper.

Figure~\ref{tbl:real_data-pl} summarizes the results for the average effort and empirical forgetting rate for the most common recall patterns. Here, we report the ratio between
the effort and empirical forgetting rate values achieved by the uniform and threshold schedules and the values achieved by \memorize with the power-law forgetting curve model.
We find that both in terms of effort and empirical forgetting rate, \memorize{} outperforms the baselines for all recall patterns. 
For example, for the recall pattern consisting of two unsuccessful recalls followed by two successful recalls (red-red-green-green), \memorize{} achieves $0.06$$\times$
lower effort and $0.11$$\times$ lower empirical forgetting rate than the second competitor.
Figure~\ref{fig:real_data_review-pl} summarizes the results for the average effort for sequences with different number of reviews, which show that \memorize{} with the power-law forgetting
curve model also outperforms the baselines.

Overall, we would like to highlight that we did not find a clear winner in performance between \memorize{} with the exponential forgetting curve model and \memorize{} with the power-law
forgetting curve model.

\section{Our Modeling Framework Using the Multiscale Context Model}\label{app:other-sde}

In this section, we will briefly describe the Multiscale Context Model (MCM) of memory~\cite{pashler2009predicting} and sketch how derive the optimal reviewing schedule for this model.

MCM models memory as \emph{activation} of $M$ time-varying \emph{context} units with the values $\{ x_{i}(t) \}_{[M]}$.
The units are \emph{leaky} and lose their activation at different decay rates $\{ \tau_i \}_{[M]}; \tau_{i + 1} > \tau_i$, such that $\Delta x_i(t + \Delta t) = x_i(t) \exp{(-\Delta t / \tau_i)}$.
The accumulated activity of the units, each scaled by $\{ \gamma_i \}_{[M]}$, represents the total \emph{trace strength} of the item at time $t$, \ie,  $s_M(t) = \frac{\sum_{i = 0}^{M} \gamma_i x_i(t)}{\sum_{i = 0}^{M} \gamma_i}$.
The probability of recall of the item at time $t$ is then calculated as $m_{MCM}(t) = \min{ \{ 1, s_M(t) \} }$.

Each time a review happens, the activation of the context pools \emph{jumps} by an amount which depends on how much activation at timescale $i$, \ie, $x_i(t)$ had \emph{contributed} to the recall.
So if a review happens at time $t_r$, $\Delta x_i(t_r^{+}) = \varepsilon(1 - s_i(t_r^{-}))$, where $s_i(t)$ represents the strength of all activations which decay \emph{faster} than the current trace, \ie, $s_i(t) = \frac{\sum_{j = 1}^{i} \gamma_j x_j(t)}{\sum_{j = 1}^{i} \gamma_j}$.
The value of $\varepsilon$ is set to $\varepsilon_{\alpha}$ if the recall was successful and to $\varepsilon_{\beta}$ if the recall was unsuccessful.
For a more detailed description of MCM, we refer the readers to~\cite{pashler2009predicting}.

The dynamics of the context pools $\{ x_i(t) \}_{[M]}$ can be converted to the corresponding SDEs readily as follows.
\begin{align}
  s_i(t) &= \frac{\sum_{j = 1}^{i} \gamma_j x_j(t)}{\sum_{j = 1}^{i} \gamma_j} \nonumber\\
  dx_{i}(t) &= \varepsilon_{\alpha}r(t)(1 - s_{i}(t))dN(t) + \varepsilon_{\beta}(1 - r(t)) (1 - s_{i}(t)) dN(t) - \frac{x_i(t)}{\tau_i}dt \label{eq:mcm-dx}\\
  ds_{M}(t) &= \frac{\sum_{j = 1}^{M} \gamma_j dx_j(t)}{\sum_{j = 1}^{M} \gamma_j}.\nonumber
\end{align}

For modeling the probability of recall $m_{MCM}(t)$, we can use a differentiable approximation to the $\min{ \{1, s_M(t)\} }$ function. For example, we can use hyperbolic-tan, and approximate $m_{MCM}(t)$ via $\tilde{m}_{MCM}(t)$:
%
%
\begin{align}
  \tilde{m}_{MCM}(t) =& \tanh{s_M(t)} \nonumber\\
  \implies d\tilde{m}_{MCM}(t) =& (1 - \tilde{m}_{MCM}^2(t)) ds_M(t).
  \label{eq:mcm-dm}
\end{align}
One can contrast Eq.~\ref{eq:mcm-dx} and Eq.~\ref{eq:mcm-dm} with Eq.~\ref{eq:forgetting-rate} and Eq.~\ref{eq:recall-probability-m} (or Eq.~\ref{eq:recall-probability-dm-pl}) respectively to compare the derivations
for the exponential forgetting curve model and the MCM.

Extension of Lemma~\ref{lem:hjb-proposals} for Eq.~\ref{eq:mcm-dm} is straight-forward and the nonlinear partial differential equation corresponding to Eq.~\ref{eq:hjb} (or Eq.~\ref{eq:hjb-pl}) can be solved to arrive at the optimal scheduling for the MCM model.
The resulting equation, however, may not readily admit to an analytical solution as was the case for the exponential and power-law forgetting curve models.

%

\section{Distribution of likelihood values for different reviewing schedules}\label{app:empirical-likelihood}

\begin{figure*}[t]
  \centering
  \subfloat[\memorize]{\includegraphics[width=0.25\textwidth]{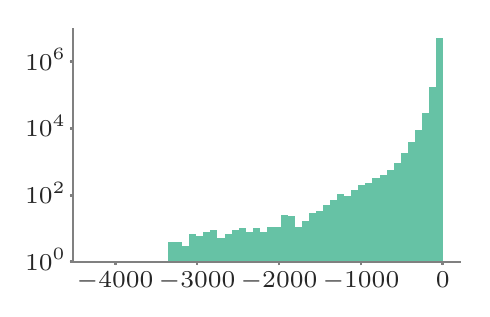}}
  \subfloat[threshold]{\includegraphics[width=0.25\textwidth]{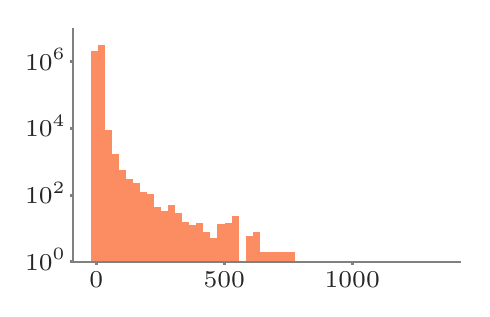}}
  \subfloat[uniform]{\includegraphics[width=0.25\textwidth]{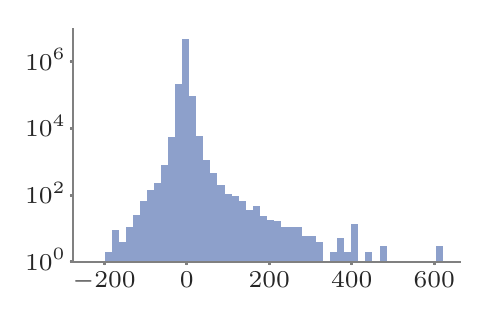}}
  \caption{Empirical distribution of log-likelihood values for all (user, item) pairs under different reviewing schedules.
        Since Duolingo uses a near-optimal hand-tuned reviewing schedule, the peak of the distribution for \memorize{} corresponds to the highest likelihood values, \ie, there are 
        many (user, item) pairs who follow \memorize{} closely.
        }
       \label{fig:likelihoods}
\end{figure*}

In this section, we compute the likelihood of each sequence of review events in our dataset under different reviewing schedules. Figure~\ref{fig:likelihoods} summarizes the results by showing the empirical distribution 
of estimated likelihood values for \memorize, threshold and uniform schedules. Since Duolingo uses a near-optimal hand-tuned reviewing schedule, the peak of the distribution for \memorize{} corresponds to the highest
likelihood values,  \ie, there are many (user, item) pairs who follow \memorize{} closely.
%

\section{Predictive performance of the memory model}\label{app:empirical-evaluation}
Before we evaluate the predictive performance of the exponential and power-law forgetting curve models, whose forgetting rates we estimated using a variant of Half-life regression (HLR)~\cite{settlestrainable},
we highlight the differences between the original HLR and the variant we used.

The original HLR and the variant we used differ in the way successful and unsuccessful recalls change the forgetting rate. In our work, the forgetting rate at time $t$ depends on $n_{\cmark}(t)=\int_0^t r(\tau)dN(\tau)$
and $n_{\xmark}(t)=\int_0^t (1-r(\tau))dN(\tau)$. In contrast, in the original HLR, the forgetting rate at time $t$ depends on $\sqrt{n_{\cmark}(t) + 1}$ and $\sqrt{n_{\xmark}(t) + 1}$.
The rationale behind our modeling choice is to be able to express the dynamics of the forgetting rate using a linear stochastic differential equation with jumps.
Moreover, Settles \ea{} consider each session to contain multiple review events for each item. Hence, within a session, the $n_{\xmark}(t)$ and $n_{\cmark}(t)$ may increase by
more than one.
In contrast, we consider each session to contain a single review event for each item because the reviews in each session take place in a very short time and it is likely that after
the first review, the user will recall the item correctly during that session. 
Hence, we only increase one of $n_{\cmark}(t)$ or $n_{\xmark}(t)$ by exactly $1$ after each session and consider an item has been successfully recalled during a session if all reviews
were successful, \ie, $p_{recall} = 1$. Noticeably, $\sim$$83\%$ of the items were successfully recalled during a session.

Table~\ref{tab:empirical-evaluation} summarizes our results on the Duolingo dataset in terms of mean absolute error (MAE), area under curve (AUC) and correlation (COR$_h$),
which show that the performance of both the exponential and power-law forgetting curve models with forgetting rates estimated using the variant of HLR is comparable to the performance
of the exponential forgetting curve model with forgetting rates estimated using the original HLR.
\begin{table}[t]
   \centering
   \begin{tabular}{r|c|c|c}
      & \multicolumn{1}{c}{\textbf{HLR}} & \multicolumn{1}{|c}{\textbf{Our Model}} & \multicolumn{1}{|c}{\textbf{Our Model}} \\
      & \multicolumn{1}{c}{Exponential} & \multicolumn{1}{|c}{Exponential} & \multicolumn{1}{|c}{Power-law} \\\hline
      MAE$\downarrow$ & 0.128 & 0.129 & \textbf{0.105}\\
      AUC$\uparrow$ & 0.538 & \textbf{0.542} & 0.533\\
      COR${}_h\uparrow$ & \textbf{0.201} & 0.165 & 0.123\\
   \end{tabular}
   \caption{Predictive performance of the exponential and power-law forgetting curve models in comparison with the results reported by
   Settles et al.~\cite{settlestrainable}. The arrows indicate whether a higher
   value of the metric is better ($\uparrow$) or a lower value ($\downarrow$).} 
   \label{tab:empirical-evaluation}
\end{table}

\section{Effect of review time on forgetting rate}\label{app:reviewTimes}

In previous studies, it has been shown that the interval between reviews has an effect on the forgetting rate, especially at large review/retention time-scales~\cite{cepeda2008spacing}.
In this section, we discuss how we tested for such effects in our dataset and justify our decision to employ the simpler model with constant updates to the forgetting rate, independent of the review interval.

\xhdr{Formulation} In Eq.~\ref{eq:forgetting-rate}, we have considered $\alpha$ and $\beta$ as constants, \ie, they do not vary with the review interval $t - t_r$, where $t_r$ is the time of last review.
We have dropped the subscript $i$ denoting the item for ease of exposition.
We can make a zeroth-order approximation to time varying $(\alpha, \beta)$ by allowing them to be piecewise constant for $K$ mutually exclusive and exhaustive review-time intervals $\left\{ B^{(i)} \right\}_{[K]}$.
We denote the value that $\alpha$ ($\beta$) takes in interval $B^{(i)}$ as $\alpha^{(i)}$ ($\beta^{(i)}$) and modify the forgetting rate update equation to
\begin{equation} 
  dn(t) = \underbrace{
    -\alpha^{(i)} n(t) r(t) dN(t) + \beta^{(i)} n(t)(1-r(t)) dN(t)
      }_{i\,  \text{ such that } t - t_r \in B^{(i)}}\nonumber
\end{equation}
If we find that $\exists\, \{ i, j \} \subset [K]$ such that $\alpha^{(i)}$ ($\beta^{(i)}$) is \emph{significantly} different from $\alpha^{(j)}$ ($\beta^{(j)}$), then we would conclude that $\alpha$ ($\beta$) vary with review-time.

We obtain repeated estimates of $\left\{ \alpha^{(i)} \right\}_{[K]}$ and $\left\{  \beta^{(i)} \right\}_{[K]}$ by fitting our model to datasets sampled with replacement from our Duolingo dataset, \ie, via bootstrapping. The Welch's t-test is used to test if the difference in mean values of the parameters in different bins is significant.

\xhdr{Experimental setup} We set the bins boundaries by determining the $K$-quantiles of the review times in our dataset. Table~\ref{tab:bins} shows that the bin boundaries for different $K$ are quite varied and adequately cover long time-scales as well as review intervals which are short enough to capture \emph{massed practicing}. This method of binning also ensures that we have sufficient samples for accurate estimation ($\sim$$5.2e6 / K$) for all parameters.
Then we use the variant of HLR described in Appendix~\ref{app:empirical-evaluation} to fit the parameters in $400$ different datasets using bootstrapping.
The regularization parameters are determined via grid-search using a train/test dataset. 
%
%
%
We thus obtain $400$ samples of $\left\{ \alpha^{(i)} \right\}_{[K]}$ and $\left\{ \beta^{(i)} \right\}_{[K]}$ for $K \in \left\{ 3, 4, 5 \right\}$ and $i \in [K]$.
Using Welch's t-test for distributions with varying variances, we observe that the mean values of the distributions of $\left\{ \alpha^{(i)} \right\}_{[K]}$ ($\left\{ \beta^{(i)} \right\}_{[K]}$) and $\left\{ \alpha^{(j)} \right\}_{[K]}$ ($\left\{ \beta^{(j)} \right\}_{[K]}$) are not significantly different for any $\{ i, j \}$. As an example, the $p$-values obtained for $K = 5$ are shown in Table~\ref{tab:p-values}.

As discussed in Section~\ref{sec:formulation}, a possible explanation for this is that our model takes the recall of the learners at each review $r(t)$ into account to update the forgetting rate while in~\cite{cepeda2008spacing}, the
updates do not take the recall into account.
\begin{table}
\centering
\begin{tabular}{ll}
\textbf{K} & \textbf{Interval boundaries}\\\midrule
3 &  [0, 20 minutes, 2.9 days, $\infty$]\\
4 &  [0, 9 minutes, 21.5 hours, 5.2 days, $\infty$]\\
5 &  [0, 6 minutes, 1.5 hours, 1.8 days, 7.3 days, $\infty$]\\
\end{tabular}
\caption{The boundaries of intervals used to divide review times into bins based on $K$-quantiles of inter-review times in the Duolingo dataset.}
\label{tab:bins}
\end{table}
\begin{table}[t]
  \centering
  \subfloat[Comparing $\alpha^{(\text{row})}$ and $\alpha^{(\text{col})}$]{%
    \begin{tabular}{c|cccc}
      & 2 & 3 & 4 & 5\\\hline
    1 & 0.317 & 0.317 & 0.317 & 0.317 \\
    2 &       & 0.312 & 0.312 & 0.312 \\
    3 &       &       & 0.317 & 0.317 \\
    4 &       &       &       & 0.318 \\
    \end{tabular}
  }
  \subfloat[Comparing $\beta^{(\text{row})}$ and $\beta^{(\text{col})}$]{%
    \begin{tabular}{c|cccc}
      & 2 & 3 & 4 & 5\\\hline
    1 & 0.165 & 0.172 & 0.165 & 0.165 \\
    2 &       & 0.318 & 0.302 & 0.302 \\
    3 &       &       & 0.318 & 0.318 \\
    4 &       &       &       & 0.302 \\
    \end{tabular}
  }
  \caption{$p$-values obtained after using Welch's t-test for populations with different variance to reject the null hypothesis that the samples (\ie, 400 samples of $\{ \alpha^{(i)} \}$ and $\{ \beta^{(i)} \}$ for $i \in [5]$) have the same mean value. In all cases, we find no evidence to reject the null hypothesis. The results for other values of $K$ were qualitatively similar.}
  \label{tab:p-values}
  \vspace{-3mm}
\end{table}
%

\end{document}